\documentclass[a4paper,twoside,english]{IEEEtran}
\usepackage[T1]{fontenc}
\usepackage[utf8]{inputenc}
\usepackage{babel}
\usepackage{pifont}
\usepackage{url}
\usepackage{amsmath}
\usepackage{amssymb}
\usepackage{graphicx}
\usepackage[numbers]{natbib}
\usepackage[unicode=true,
 bookmarks=true,bookmarksnumbered=false,bookmarksopen=false,
 breaklinks=false,pdfborder={0 0 1},backref=false,colorlinks=false]
 {hyperref}

\makeatletter

\pdfpageheight\paperheight
\pdfpagewidth\paperwidth

\newcommand*{\LyxTextAccent}[3][0ex]{%
  \hmode@bgroup\ooalign{\null#3\crcr\hidewidth
  \raise#1\hbox{#2}\hidewidth}\egroup}
\newcommand{\LyxAccentSize}[1][\sf@size]{%
  \check@mathfonts\fontsize#1\z@\math@fontsfalse\selectfont
}
\ProvideTextCommandDefault{\textcommabelow}[1]{
  \LyxTextAccent[-.31ex]{\LyxAccentSize,}{#1}}

\providecommand{\tabularnewline}{\\}

\AtBeginDocument{

}

\makeatother

\begin{document}
\title{MambaDepth: Enhancing Long-range Dependency for Self-Supervised Fine-Structured
Monocular Depth Estimation}
\author{Ionu\textcommabelow{t} GRIGORE, Călin-Adrian POPA}
\IEEEspecialpapernotice{Department of Computers and Information Technology\\ Politehnica
University of Timi\textcommabelow{s}oara\\ Blvd. V. Pârvan, No. 2,
300223 Timi\textcommabelow{s}oara, Romania}
\maketitle
\begin{abstract}
In the field of self-supervised depth estimation, Convolutional Neural
Networks (CNNs) and Transformers have traditionally been dominant.
However, both architectures struggle with efficiently handling long-range
dependencies due to their local focus or computational demands. To
overcome this limitation, we present MambaDepth, a versatile network
tailored for self-supervised depth estimation. Drawing inspiration
from the strengths of the Mamba architecture, renowned for its adept
handling of lengthy sequences and its ability to capture global context
efficiently through a State Space Model (SSM), we introduce MambaDepth.
This innovative architecture combines the U-Net's effectiveness in
self-supervised depth estimation with the advanced capabilities of
Mamba. MambaDepth is structured around a purely Mamba-based encoder-decoder
framework, incorporating skip connections to maintain spatial information
at various levels of the network. This configuration promotes an extensive
feature learning process, enabling the capture of fine details and
broader contexts within depth maps. Furthermore, we have developed
a novel integration technique within the Mamba blocks to facilitate
uninterrupted connectivity and information flow between the encoder
and decoder components, thereby improving depth accuracy. Comprehensive
testing across the established KITTI dataset demonstrates MambaDepth's
superiority over leading CNN and Transformer-based models in self-supervised
depth estimation task, allowing it to achieve state-of-the-art performance.
Moreover, MambaDepth proves its superior generalization capacities
on other datasets such as Make3D and Cityscapes. MambaDepth's performance
heralds a new era in effective long-range dependency modeling for
self-supervised depth estimation. Code is available at \url{https://github.com/ionut-grigore99/MambaDepth}.
\end{abstract}

\section{Introduction}

Accurate depth estimation from single image is an active research
field to help computer reconstruct and understand the real scenes.
It also has a large range of applications in diverse fields such as
autonomous vehicles, robotics, augmented reality, etc. While supervised
monocular depth estimation has been successful, it is suffering from
the expensive access to ground truth. Furthermore, supervised depth
estimators often face optimization challenges under sparse supervision
and exhibit limited adaptability to new, unencountered scenarios. 

\begin{figure}[t]
\centering \includegraphics[scale=0.12]{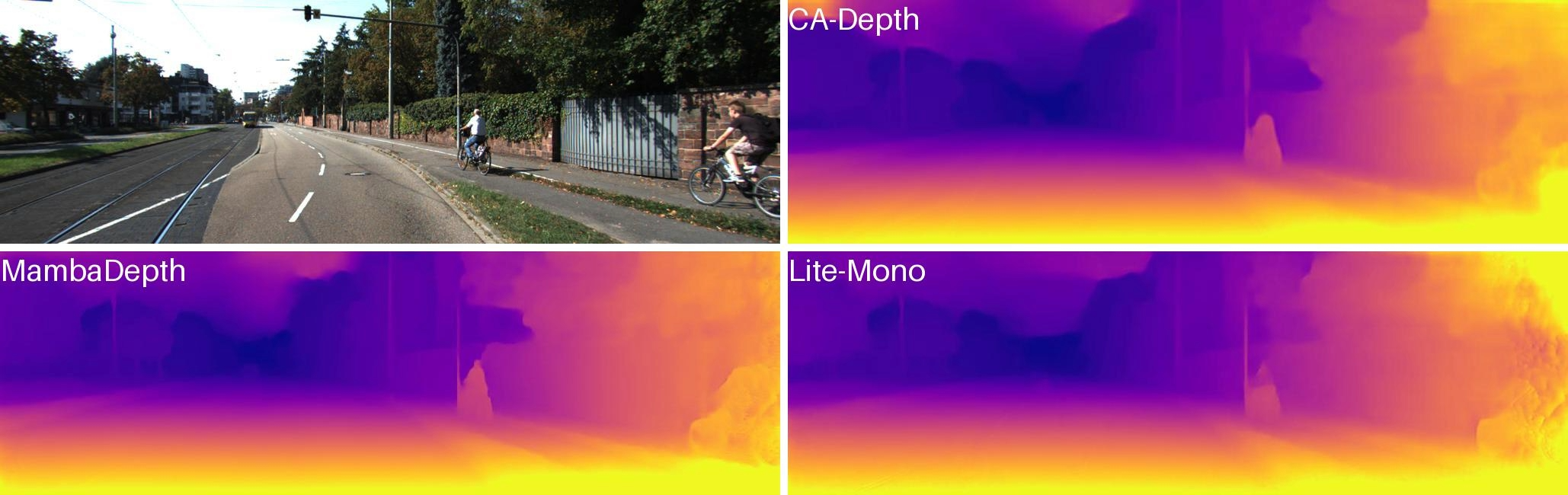}\label{fig:short-b-2-1-1}

\caption{Our method's typical predictions on images from the KITTI dataset
exhibit superior performance when compared to the classical Monodepth2
\citep{key-17} and the contemporary attempts to use Transformers
\citep{key-73} or self-attention mechanism \citep{key-74} in self-supervised
monocular depth estimation. Notably, our approach excels in recovering
intricate scene details.}
\label{fig:MambaDepth's predictions against Monodepth2 and Transformer-based and attention-based methods}
\end{figure}

Recently, self-supervised approaches have become increasingly prominent.
Current strategies primarily focus on utilizing self-distillation
techniques \citep{key-45}, incorporating depth hints \citep{key-59},
and employing multi-frame inference \citep{key-60,key-11}. Despite
these advancements, a common shortfall is their inability to capture
detailed scene intricacies, as illustrated in Figure \ref{fig:MambaDepth's predictions against Monodepth2 and Transformer-based and attention-based methods}.
The challenge lies in effectively and efficiently learning these fine-grained
structural details within a self-supervised framework.

CNNs \citep{key-36} and Transformers \citep{key-56} are two significant
architectures in the realm of self-supervised depth estimation. CNNs,
like Monodepth2 \citep{key-17} and MiDaS \citep{key-49}, excel in
hierarchical feature extraction with more efficiency in parameters
compared to traditional fully connected networks. Their weight-sharing
structure is key in identifying translational invariances and local
patterns. Conversely, Transformers, initially developed for natural
language processing, have adapted well to image processing tasks.
Examples include Vision Transformer (ViT) \citep{key-8} in image
recognition and SwinTransformer \citep{key-40} as a versatile vision
task backbone. Unlike CNNs, Transformers process images not as spatial
hierarchies but as sequences of patches, enhancing their global information
capturing ability. This distinction has led to the emergence of hybrid
architectures combining CNNs and Transformers, like Depthformer \citep{key-38},
TransDepth \citep{key-61}, and DPT \citep{key-48}. 

Transformers, while adept at managing long-range dependencies, pose
a significant computational load due to the self-attention mechanism's
quadratic scaling with input size. This is particularly challenging
for high-resolution images as ones used for depth estimation. To address
this, state space sequence models (SSMs) \citep{key-21,key-25}, especially
structured state space sequence models (S4) \citep{key-24}, have
shown promise. They offer an efficient and effective approach for
deep network construction, as seen in Mamba \citep{key-22}, which
optimizes S4 with a selective mechanism and hardware-aware design.
These models have shown potential in language, genomics, and also
in vision tasks like image \citep{key-43} and video classification
\citep{key-29}. Given that image patches and features can be treated
as sequences \citep{key-40,key-8}, this encourages the exploration
of SSMs, particularly the Mamba blocks, to enhance U-Net long-range
modeling capabilities.

Our main contributions are as follows:
\begin{itemize}
\item We propose MambaDepth, a versatile network designed for self-supervised
depth estimation, utilizing a novel SSM-based structure which effectively
captures both localized details and extensive dependencies within
images and thus obtains fine-grained scene geometry of a single image.
To the best of our knowledge, this is the first time SSMs are used
for self-supervised depth estimation. It distinguishes itself from
conventional Transformer-based models by offering linear feature size
scaling, avoiding the Transformers' usual quadratic complexity. Additionally,
we have developed a novel integration technique for Mamba blocks that
ensures seamless connectivity and information flow between the encoder
and decoder components, thereby enhancing depth accuracy.
\item Our extensive evaluations across the KITTI dataset show MambaDepth's
exceptional performance, significantly outperforming Transformer-based
networks and existing self-supervised alternatives in accuracy and
efficiency. Moreover, MambaDepth's enhanced generalization is showcased
through its application of a KITTI pre-trained model to diverse datasets,
including successful zero-shot transfer to Make3D and Cityscapes. 
\item This breakthrough sets the stage for future network designs that efficiently
and effectively handle long-range dependencies in self-supervised
depth estimation.
\end{itemize}

\section{Related work}

\subsection{Supervised Depth Estimation}

Eigen and colleagues \citep{key-10} were pioneers in adopting a learning-based
strategy, employing a multiscale convolutional neural network combined
with a scale-invariant loss function to estimate depth from a single
image. This groundbreaking approach has since inspired a plethora
of subsequent methodologies. Broadly, these techniques fall into two
main categories: one views depth estimation as a problem of pixel-wise
regression, as seen in works like \citep{key-10}, \citep{key-28},
\citep{key-48}, and \citep{key-66}. The other approach treats it
as a pixel-wise classification challenge, as demonstrated in studies
\citep{key-12} and \citep{key-7}. While regression-based methods
are capable of predicting continuous depth values, they often present
optimization challenges. On the other hand, classification-based methods,
though simpler to optimize, are limited to predicting discrete depth
values. In an innovative attempt to harness the advantages of both
regression and classification, certain studies, notably those referenced
as \citep{key-1,key-31}, have redefined depth estimation as a dual
task involving both classification and regression at the pixel level.
This method involves initially regressing a series of depth bins followed
by a pixel-wise classification, where each pixel is assigned to its
respective bin. The ultimate depth value is then derived as a linear
amalgamation of the centers of these bins, with weights given by their
respective probabilities. This hybrid technique has shown significant
enhancements in terms of accuracy.

\begin{figure*}
\centering \includegraphics[scale=0.33]{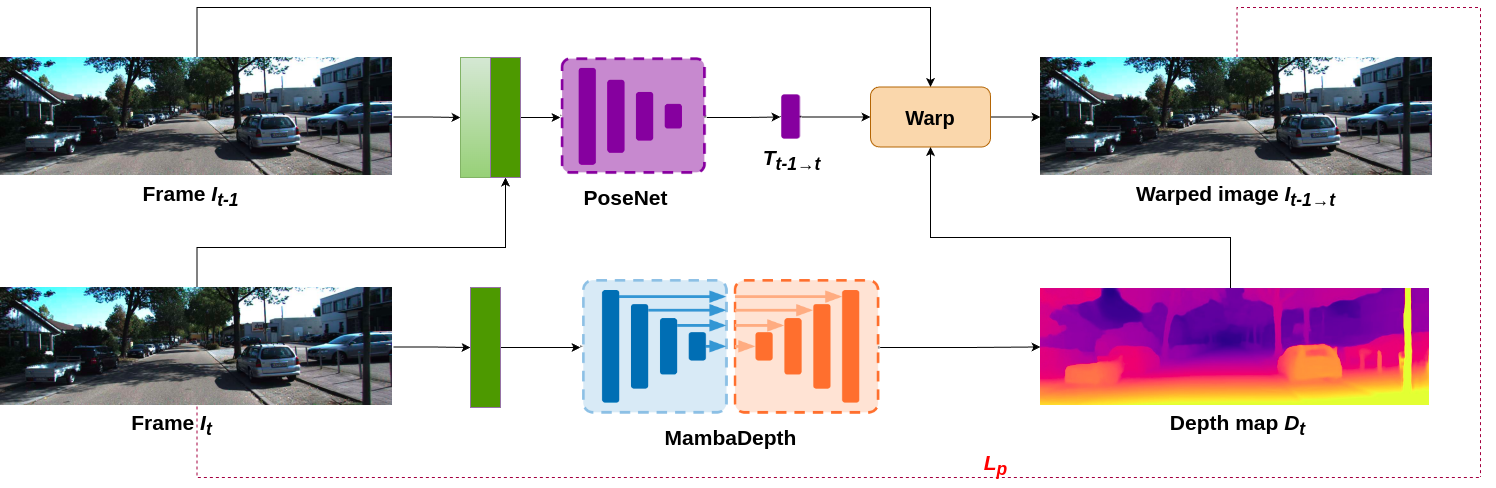}\label{fig:short-b}
\caption{\textbf{Overview of our self-supervised framework}. Our proposed MambaDepth
adopts a U-Net architecture, leveraging MambaDepth blocks from encoder
to obtain low-resolution feature maps of the current frame $I_{t}$.
Subsequently, low-resolution feature maps traverse successive MambaDepth
blocks from the decoder together with skip connections in order to
obtain disparities after applying a final Sigmoid layer. The predicted
disparities are then upsampled at various scales to match the original
input resolutions. Additionally, a standard pose network utilizes
temporally adjacent frames $I_{t}$ and $I_{t-1}$ as input, yielding
relative pose $T_{t-1\rightarrow t}$ as output. The camera pose is
solely required during training for conducting differentiable warping.
In line with numerous prior studies, we employ pixels from frame $I_{t-1}$
to reconstruct frame $I_{t}$ using the depth map $D_{t}$ and relative
pose $T_{t-1\rightarrow t}$ through a differentiable warping process
\citep{key-30}. The loss function is formulated based on the differences
between the warped image $I_{t-1\rightarrow t}$ and the source image
$I_{t}$.}
\label{fig:Overview of our self-supervised framework}
\end{figure*}

\subsection{Self-supervised Depth Estimation}

In scenarios lacking direct ground truth data, self-supervised models
are often developed by leveraging either the temporal consistency
found in monocular video sequences, as explored in studies \citep{key-17,key-67},
or the left-right consistency observed in stereo image pairs, a concept
investigated in references \citep{key-46,key-14,key-16}.

\textbf{Monocular Training Approach}: This method derives supervision
from the congruence between a synthesized scene view from a reference
frame and the actual view from a source frame. A notable example,
the SfMLearner \citep{key-46}, synchronizes the training of a DepthNet
and a separate PoseNet using a photometric loss function. Building
on this foundational approach, numerous enhancements have been proposed.
These include robust image-level reconstruction losses \citep{key-53,key-20},
feature-level reconstruction losses \citep{key-53,key-64}, incorporation
of auxiliary information during training \citep{key-34,key-59}, strategies
to address the dynamic objects that disrupt static scene assumptions
\citep{key-17,key-20,key-50,key-57,key-5,key-2,key-3,key-33,key-37,key-63},
and additional constraints \citep{key-62,key-63,key-50,key-4,key-18,key-26,key-69}.

\textbf{Stereo Training Method}: Here, synchronized stereo image pairs
are used, focusing on predicting a disparity map \citep{key-52},
which is effectively the inverse of a depth map. With known relative
camera poses, the model's task is simplified to disparity map prediction.
Garg et al. \citep{key-14} pioneered this with a self-supervised
monocular depth estimator, applying a photometric consistency loss
between stereo pairs. Subsequent improvements include implementing
left-right consistency \citep{key-16} and temporal consistency in
videos \citep{key-64}. Garg et al. \citep{key-13} further refined
this by enabling the prediction of continuous disparity values. Stereo-based
methods have evolved to include semi-supervised data \citep{key-35,key-41},
auxiliary information usage \citep{key-59}, exponential probability
volumes \citep{key-19}, and self-distillation techniques \citep{key-47,key-27,key-45}.
Stereo views offer an ideal reference for supervision and can also
be instrumental in deriving absolute depth scales.

However, existing self-supervised methods still struggle with producing
high-fidelity depth maps. Current techniques primarily rely on immediate
visual features or utilize Transformer \citep{key-8} enhanced high-level
visual representations, often neglecting the critical role of pixel-level
geometric cues that could significantly enhance model performance
and generalization abilities.

\subsection{State Space Models}

State Space Sequence Models (SSMs) \citep{key-21} represent a category
of systems that transform a one-dimensional function or sequence $u(t)$
into $y(t)$. They are described by the following linear Ordinary
Differential Equation (ODE):
\begin{equation}
x'(t)=Ax(t)+Bu(t),\label{eq:important}
\end{equation}
\begin{equation}
y(t)=Cx(t)+Du(t).\label{eq:also-important}
\end{equation}

In this equation, $A\in\mathbb{R}^{N\times N}$ is the state matrix,
and $B,C\in\mathbb{R}^{N}$ are its parameters and skip connection
$D\in\mathbb{R}^{1}$, with $x(t)\in\mathbb{R}^{N}$symbolizing the
implicit latent state. SSMs possess several advantageous properties,
such as linear complexity per time step and the ability for parallel
computations, which aid in efficient training. Yet, standard SSMs
generally require more memory compared to equivalent CNNs and face
issues like vanishing gradients during training, limiting their widespread
use in sequence modeling.

The evolution of SSMs led to the creation of Structured State Space
Sequence Models (S4) \citep{key-24}, which notably enhance the basic
SSM framework. S4 achieves this by applying structured designs to
the state matrix $A$ and incorporating an efficient algorithm. The
state matrix in S4 is specifically developed and initialized using
the High-Order Polynomial Projection Operator (HIPPO) \citep{key-23},
facilitating the construction of deep sequence models that are both
rich in capability and adept at long-range reasoning. Remarkably,
S4 has outperformed Transformers \citep{key-56} in the demanding
Long Range Arena Benchmark \citep{key-54}.

Mamba \citep{key-22} represents a further advancement in SSMs, especially
in discrete data modeling, such as text and genomic sequences. Mamba
introduces two significant enhancements. Firstly, it incorporates
an input-specific selection mechanism, differing from traditional,
invariant SSMs. This mechanism filters information efficiently by
customizing SSM parameters based on input data. Secondly, Mamba employs
a hardware-optimized algorithm, which scales linearly with sequence
length and uses a scanning process for recurrent computation, enhancing
speed on contemporary hardware. Mamba's architecture, which combines
SSM blocks with linear layers, is notably more streamlined. It has
achieved top-tier results in various long-sequence fields, including
language and genomics, demonstrating considerable computational efficiency
in both training and inference phases.

\section{Method}

\begin{figure*}
\centering \includegraphics[scale=0.29]{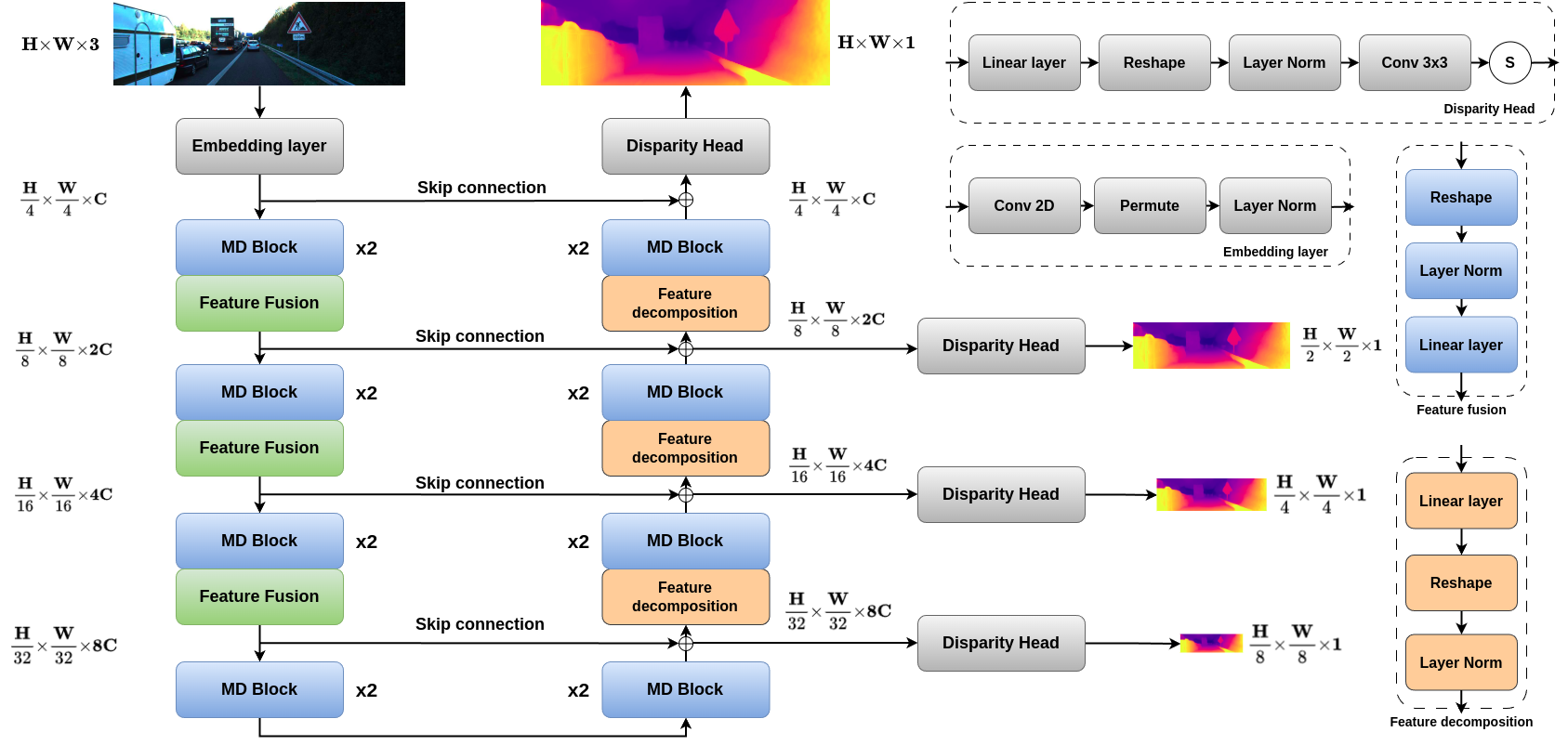}\label{fig:short-b-2}
\caption{\textbf{Overview of MambaDepth architecture}. The MambaDepth structure
includes an encoder, bottleneck, a decoder, and skip connections.
Each of these components -- the encoder, bottleneck, and decoder
-- is built using the MD block.}
\label{fig:Overview of MambaDepth architecture}
\end{figure*}

\subsection{Self-supervised framework}

In this section we describe the framework of our model and describe
how we provide the supervisory signal during the training of our model.
Fundamentally, our method is a form of Structure from Motion (SfM),
where the monocular camera is moving within a rigid environment to
provide multiple views of that scene. Our framework is built upon
Monodepth2 \citep{key-17}.

Let $I_{t}\in\mathbb{R}^{H\times W\times3},t\in\left\{ -1,0,1\right\} $
be a frame in a monocular video sequence captured by a moving camera,
where $t$ is the frame time index. Similarly, let $D_{t}\in\mathbb{R}^{H\times W}$
denote the depth map corresponding to image $I_{t}$. The camera pose
changes from time $0$ to time $t,t\in\left\{ -1,1\right\} $ is encoded
by the $3\times3$ rotation matrix $R_{t}$ and the translation vector
$t_{t}$. We obtain the $4\times4$ camera transformation matrix thus:
\begin{equation}
M_{t}=\begin{bmatrix}R_{t} & t_{t}\\
0 & 1
\end{bmatrix}.\label{eq:important-2}
\end{equation}
Our aim is to train two CNN networks to simultaneously estimate the
pose of the camera, and the structure of the scene, respectively:
\begin{equation}
M=\theta_{pose}(I_{t}),\label{eq:important-1}
\end{equation}
\begin{equation}
D=\theta_{depth}(I_{t}).\label{eq:also-important-1}
\end{equation}
Self-supervised depth prediction reformulates the learning task as
a novel view-synthesis problem. Specifically, during training, we
let the coupled network synthesize the photo-consistency appearance
of a target frame from another viewpoint of the source frame. We treat
the depth map as an intermediate variable to constrain the network
to complete the image synthesis task. 

Let $(u,v)\in\mathbb{R}^{2}$ be the calibrated coordinates of a pixel
in image $I_{0}$. In this case, let the origin $(0,0)$ be the top-left
of the image. In the process of imaging, a 3D point $(X,Y,Z)\in\mathbb{R}^{3}$
projects onto $(u,v)$ through a perspective projection operator. 

Suppose that the transformation matrix $M_{t}$ encodes the pose change
of the camera from time $0$ to time $t$ and \eqref{eq:important-3}
is the perspective projection operator:
\begin{equation}
\pi(X,Y,Z)=\left(f_{x}\frac{X}{Z}+c_{x},f_{y}\frac{Y}{Z}+c_{y}\right)=(u,v),\label{eq:important-3}
\end{equation}
where $(f_{x},f_{y},c_{x},c_{y})$ are the camera intrinsic parameters.
Therefore, given a depth map $D(u,v)$, a 2D image point $(u,v)$
backprojects to a 3D point $(X,Y,Z)$ through backprojection operator:
\begin{equation}
\pi^{-1}(u,v,D(u,v))=D(u,v)\left(\frac{u-c_{x}}{f_{x}},\frac{v-c_{y}}{f_{y}}\right)=(X,Y,Z).\label{eq:important-4}
\end{equation}
Then the corresponding pixels in image $I_{t}$ can be computed as:
\begin{equation}
(u',v')=\pi(M_{t}\pi^{-1}(u,v,D(u,v)))=g(u,v|D(u,v),M_{t}).\label{eq:important-5}
\end{equation}
We project the pixels of an image to form a novel synthetic view \eqref{eq:important-5}.
However, the projected coordinates $(u',v')$ are continuous values.
To obtain $I_{s}(u,v)$ we include a differentiable bilinear sampling
mechanism, as proposed in spatial transformer networks \citep{key-30}.
We can now linearly interpolate the values of the 4-pixel neighbors
(top-left, top-right, bottom-left, bottom-right) of $I(u',v')$ to
give the RGB intensities as follows:
\begin{equation}
I^{s}(u,v)=\sum_{u}\sum_{v}w^{uv}I(u',v'),\label{eq:important-6}
\end{equation}
where $w^{uv}$ is linearly proportional to the spatial proximity
between $(u,v)$ and $(u',v')$, and $\sum_{u,v}w^{uv}=1$. 

\begin{figure*}
\centering \includegraphics[scale=0.11]{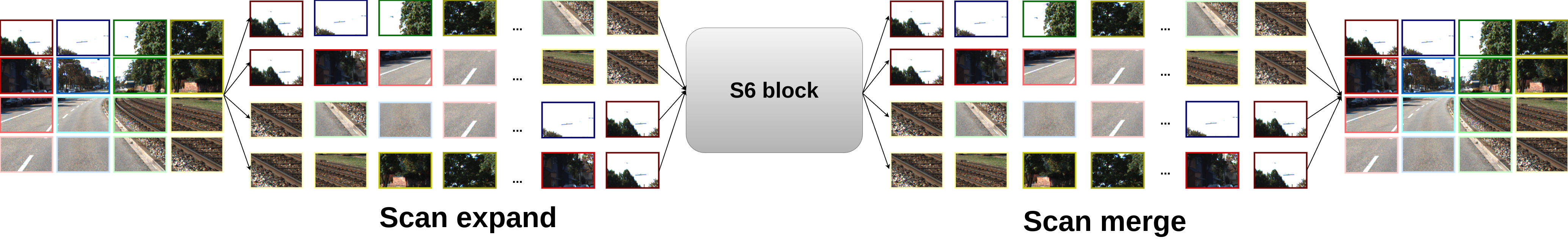}\label{fig:short-b-3-1}
\caption{The scan expanding and scan merging operations in SS2D. In the SS2D
method, input patches follow four distinct scanning paths. Each sequence
is then independently processed by separate S6 blocks. Finally, the
results are combined to create a 2D feature map, which serves as the
final output.}
\label{fig:The scan expanding and scan merging operations in SS2D}
\end{figure*}

\subsection{MambaDepth}

In this section, we elaborate the design details of the core components
in the MambaDepth.

The Mamba framework has shown remarkable efficacy in processing various
kinds of discrete data. However, its application in image data processing,
especially in the field of self-supervised depth estimation, has not
been fully explored. \citep{key-42} proposes U-Mamba, a hybrid CNN-SSM
architecture, to handle the long-range dependencies in biomedical
image segmentation and \citep{key-68} build a pure SSM-based model,
which can be adopted as a generic vision backbone, but their efficiency
are not yet fully understood at a large scale. Images, fundamentally
discrete samples from continuous signals, can be transformed into
extended sequences. This characteristic suggests the potential of
using Mamba's linear scaling benefits to improve the capability of
UNet architecture in modeling extensive range dependencies. While
image processing using Transformers, like ViT and SwinTransformer,
has seen success, their application is limited by the significant
computational demands for large images due to the self-attention mechanism's
quadratic complexity. This challenge presents an opportunity to utilize
Mamba's linear scaling to bolster UNet capacity for long-range dependency
modeling.

In self-supervised depth estimation, Monodepth2 \citep{key-17} and
its derivatives, recognized for their symmetric encoder-decoder structure,
are predominant. This structure is adept at extracting multi-level
image features through convolutional methods. However, the design
is constrained in its ability to capture long-range dependencies in
images, as the convolutional kernels focus on local areas. Each convolutional
layer only processes features within its limited receptive field.
Although skip connections aid in merging detailed and abstract features,
they primarily enhance local feature combination, not the modeling
of extensive range dependencies.

MambaDepth, a novel design, integrates the strengths of Monodepth2
\citep{key-17} and Mamba \citep{key-22} to comprehensively understand
global contexts in self-supervised depth estimation. Figure \ref{fig:Overview of MambaDepth architecture}
showcases the structure of MambaDepth, which is distinct in its composition,
featuring an embedding layer, encoder, decoder, disparity heads, and
straightforward skip connections, marking a departure from the classical
designs often found in prior work.

In the initial stage, the embedding layer segments the input image,
denoted as $x$ with dimensions $\mathbb{R}^{H\times W\times3}$,
into discrete $4\times4$ patches. These patches are then transformed
to a predefined dimension $C$ (typically set to 96), resulting in
a reshaped image representation, $x'$, with dimensions $\mathbb{R}^{\frac{H}{4}\times\frac{W}{4}\times C}$.
$x'$ is normalized using Layer Normalization before its progression
into the encoder for the extraction of features. The encoder itself
is structured into four phases, incorporating our newly introduced
feature fusion step at the conclusion of the first three phases to
compact the spatial dimensions and amplify the channel depth, employing
a configuration of $[2,2,2,2]$ MD blocks and channel dimensions scaling
from $[C,2C,4C,8C]$ through each phase.

Conversely, the decoder reverses this process over its four stages,
using our introduced feature decomposition technique at the start
of the final three stages to enlarge spatial dimensions and condense
the channel count. Here, the arrangement of MD blocks is $[2,2,2,2]$,
with channel dimensions inversely scaling from $[8C,4C,2C,C]$. The
decoder culminates in 4 disparity heads that upscale the feature dimensions
by a factor of four through feature decomposition, followed by a projection
layer that adjusts the channel count to align with the target of self-supervised
depth estimation, which is processed through a convolutional layer
and a Sigmoid layer to generate the final depth map. 

Skip connections within this architecture are implemented via a simple
addition operation, purposely designed to avoid the incorporation
of extra parameters, thus maintaining the model's efficiency and simplicity. 

\subsection{MD block}

\begin{figure}[t]
\centering \includegraphics[scale=0.33]{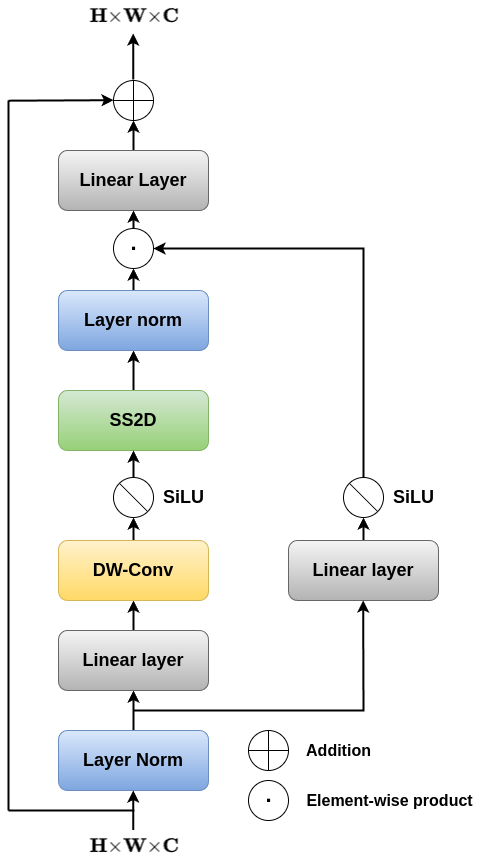}\label{fig:short-b-3-2}

\caption{The detailed structure of the MD (MambaDepth) Block.}
\label{fig:The detailed structure of the MD Block}
\end{figure}

\begin{table*}
\centering %
\begin{tabular}{|c|c|c|c|c|c|c|c|c|c|c|}
\hline 
Method & Train & Params & H$\times$W & AbsRel $\downarrow$ & SqRel $\downarrow$ & RMSE $\downarrow$ & RMSESlog $\downarrow$ & $\delta<1.25$$\uparrow$ & $\delta<1.25^{2}$$\uparrow$ & $\delta<1.25^{3}$$\uparrow$\tabularnewline
\hline 
\hline 
MonoDepth2 \citep{key-17} & M & 34M & 192$\times$640 & 0.110 & 0.831 & 4.642 & 0.187 & 0.883 & 0.962 & 0.982\tabularnewline
CaDepth-Net \citep{key-74} & M & 58M & 192$\times$640 & 0.105 & 0.769 & 4.535 & 0.181 & 0.892 & 0.964 & 0.983\tabularnewline
PackNet-SfM \citep{key-78} & M & 128M & 192$\times$640 & 0.107 & 0.785 & 4.612 & 0.185 & 0.887 & 0.962 & 0.982\tabularnewline
HR-Depth \citep{key-72} & M & 14M & 192$\times$640 & 0.109 & 0.792 & 4.632 & 0.185 & 0.884 & 0.962 & 0.983\tabularnewline
Lite-Mono \citep{key-73} & M & 8M & 192$\times$640 & 0.107 & 0.765 & 4.561 & 0.183 & 0.886 & 0.963 & 0.983\tabularnewline
DynamicDepth \citep{key-11} & M & - & 192$\times$640 & \textbf{0.096} & 0.720 & 4.458 & 0.175 & 0.897 & 0.964 & 0.984\tabularnewline
MonoViT \citep{key-65} & M & 27M & 192$\times$640 & 0.099 & 0.708 & 4.372 & 0.175 & 0.900 & 0.967 & 0.984\tabularnewline
MambaDepth (Ours) & M & 30M & 192$\times$640 & 0.097 & \textbf{0.706} & \textbf{4.370} & \textbf{0.172} & \textbf{0.907} & \textbf{0.970} & \textbf{0.986}\tabularnewline
\hline 
MonoDepth2 \citep{key-17} & MS & 34M & 320$\times$1024 & 0.106 & 0.806 & 4.630 & 0.193 & 0.876 & 0.958 & 0.980\tabularnewline
MonoDepth2 \citep{key-17} & M & 34M & 320$\times$1024 & 0.115 & 0.882 & 4.701 & 0.190 & 0.879 & 0.961 & 0.982\tabularnewline
HR-Depth \citep{key-72} & MS & 14M & 320$\times$1024 & 0.101 & 0.716 & 4.395 & 0.179 & 0.899 & 0.966 & 0.983\tabularnewline
HR-Depth \citep{key-72} & M & 14M & 320$\times$1024 & 0.106 & 0.755 & 4.472 & 0.181 & 0.892 & 0.966 & 0.984\tabularnewline
EPCDepth \citep{key-45} & S & - & 320$\times$1024 & 0.091 & 0.646 & 4.207 & 0.176 & 0.901 & 0.966 & 0.983\tabularnewline
Depth Hints \citep{key-59} & MS & - & 320$\times$1024 & 0.098 & 0.702 & 4.398 & 0.183 & 0.887 & 0.963 & 0.983\tabularnewline
Depth Hints \citep{key-59} & S & - & 320$\times$1024 & 0.096 & 0.710 & 4.393 & 0.185 & 0.890 & 0.962 & 0.981\tabularnewline
CADepth-Net \citep{key-74} & MS & 58M & 320$\times$1024 & 0.096 & 0.964 & 4.264 & 0.173 & 0.908 & \textbf{0.968} & 0.984\tabularnewline
CADepth-Net \citep{key-74} & M & 58M & 320$\times$1024 & 0.102 & 0.734 & 4.407 & 0.178 & 0.898 & 0.966 & 0.984\tabularnewline
MT-SfMLearner \citep{key-79} & M & - & 320$\times$1024 & 0.104 & 0.799 & 4.547 & 0.181 & 0.893 & 0.963 & 0.982\tabularnewline
Lite-Mono \citep{key-73} & M & 8M & 320$\times$1024 & 0.097 & 0.710 & 4.309 & 0.174 & 0.905 & 0.967 & 0.984\tabularnewline
MonoViT \citep{key-65} & M & 27M & 320$\times$1024 & 0.096 & 0.714 & 4.292 & 0.172 & 0.908 & \textbf{0.968} & 0.984\tabularnewline
MambaDepth (Ours) & M & 30M & 320$\times$1024 & \textbf{0.095} & \textbf{0.634} & \textbf{3.402} & \textbf{0.169} & \textbf{0.914} & \textbf{0.968} & \textbf{0.985}\tabularnewline
\hline 
\end{tabular}

\caption{\textbf{Performance comparison on KITTI} \citep{key-15} \textbf{eigen
benchmark}. Best results are marked in bold.}
\label{tab:Performance comparison on KITTI eigen benchmark}
\end{table*}

\begin{table*}
\centering %
\begin{tabular}{|c|c|c|c|c|c|c|c|c|c|c|}
\hline 
Method & Train & Params & H$\times$W & AbsRel $\downarrow$ & SqRel $\downarrow$ & RMSE $\downarrow$ & RMSESlog $\downarrow$ & $\delta<1.25$$\uparrow$ & $\delta<1.25^{2}$$\uparrow$ & $\delta<1.25^{3}$$\uparrow$\tabularnewline
\hline 
\hline 
MonoDepth2 \citep{key-17} & M & 34M & 192$\times$640 & 0.090 & 0.545 & 3.942 & 0.137 & 0.914 & 0.983 & 0.995\tabularnewline
HR-Depth \citep{key-72} & M & 14M & 192$\times$640 & 0.085 & 0.471 & 3.769 & 0.130 & 0.919 & 0.985 & 0.996\tabularnewline
PackNet-SfM \citep{key-78} & M & 128M & 192$\times$640 & 0.078 & 0.420 & 3.485 & 0.121 & 0.931 & 0.986 & 0.996\tabularnewline
CADepth-Net \citep{key-74} & M & 58M & 192$\times$640 & 0.080 & 0.450 & 3.649 & 0.124 & 0.927 & 0.986 & 0.996\tabularnewline
DIFFNet \citep{key-77} & M & - & 192$\times$640 & 0.076 & 0.412 & 3.494 & 0.119 & 0.935 & 0.988 & 0.996\tabularnewline
MonoViT \citep{key-65} & M & 27M & 192$\times$640 & 0.075 & 0.389 & 3.419 & 0.115 & 0.938 & 0.989 & \textbf{0.997}\tabularnewline
MambaDepth (Ours) & M & 30M & 192$\times$640 & \textbf{0.073} & \textbf{0.386} & \textbf{3.415} & \textbf{0.112} & \textbf{0.941} & \textbf{0.991} & \textbf{0.997}\tabularnewline
\hline 
MonoDepth2 \citep{key-17} & MS & 34M & 320$\times$1024 & 0.091 & 0.531 & 3.742 & 0.135 & 0.916 & 0.984 & 0.995\tabularnewline
MonoDepth2 \citep{key-17} & S & 34M & 320$\times$1024 & 0.077 & 4.455 & 3.489 & 0.119 & 0.933 & 0.988 & 0.996\tabularnewline
DepthHints \citep{key-59} & S & - & 320$\times$1024 & 0.074 & \textbf{0.364} & \textbf{3.202} & \textbf{0.114} & 0.936 & \textbf{0.989} & \textbf{0.997}\tabularnewline
MambaDepth (Ours) & M & 30M & 320$\times$1024 & \textbf{0.072} & 0.410 & 3.490 & \textbf{0.114} & \textbf{0.940} & \textbf{0.989} & 0.996\tabularnewline
\hline 
\end{tabular}

\caption{\textbf{Performance comparison using KITTI} \textbf{improved ground
truth} \citep{key-55}. Best results are marked in bold.}
\label{tab:Performance comparison using KITTI improved ground truth}
\end{table*}

At the heart of MambaDepth lies the MD (Mamba Depth) module, which
is an adaptation from VMamaba \citep{key-39}, as illustrated in Figure
\ref{fig:The detailed structure of the MD Block}. This module begins
with Layer Normalization of the input, which then bifurcates into
two distinct pathways. The initial pathway channels the input through
a linear transformation and an activation phase, whereas the second
pathway subjects the input to a sequence involving a linear transformation,
depthwise separable convolution, and activation, before directing
it to the 2D-Selective-Scan (SS2D) component for advanced feature
extraction. Following this, the extracted features undergo Layer Normalization,
are then merged with the output of the first pathway through element-wise
multiplication, and finally are integrated using a linear transformation.
This process is augmented with a residual connection, culminating
in the output of the MD block. For activation, the SiLU function is
consistently utilized throughout this study.

The SS2D mechanism comprises three essential stages: an operation
to expand the scan, an S6 block for processing, and a merging operation
for the scans. As shown in Figure \ref{fig:The scan expanding and scan merging operations in SS2D},
the expansion phase unfolds the input image in four orientations (diagonally
and anti-diagonally) into sequences. These sequences are then refined
by the S6 block, a procedure that meticulously scans information from
all directions to extract a comprehensive range of features. Following
this, the sequences are recombined through a merging operation, ensuring
the output image is resized back to the original dimensions. The innovative
S6 block, evolving from Mamba and building upon the S4 structure,
introduces a selective filter that dynamically adjusts to the input
by fine-tuning the parameters of the State Space Model (SSM). This
adjustment allows the system to selectively focus on and preserve
relevant information, while discarding what is unnecessary. 

\begin{figure*}
\centering \includegraphics[scale=0.16]{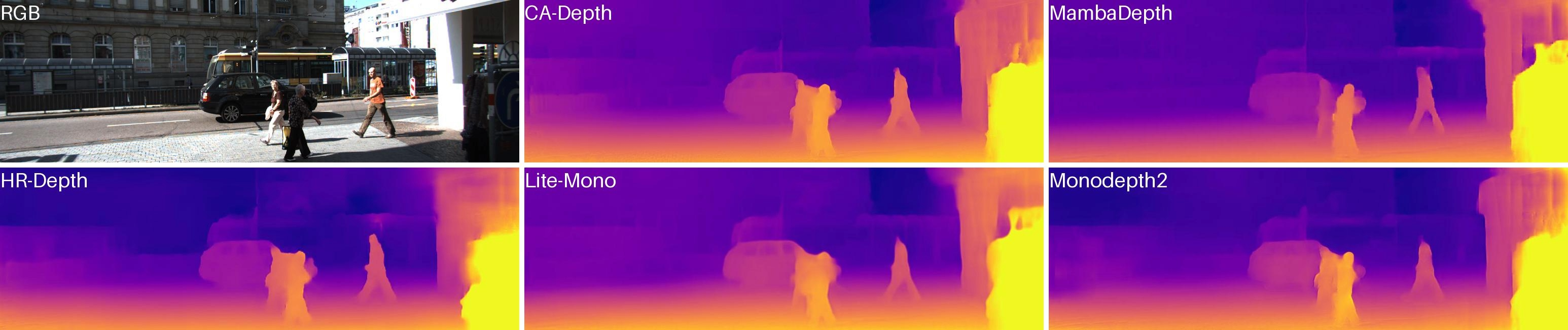}\label{fig:short-b-2-1}
\caption{\textbf{Qualitative results on the KITTI eigen benchmark. }}

\label{fig:Qualitative results on the KITTI eigen benchmark}
\end{figure*}

\subsection{Loss function}

\textbf{Objective functions.} In line with the methodologies described
in \citep{key-16,key-17}, we adopt the conventional photometric loss
$pe$, which is a combination of $L1$ and $SSIM$ losses:
\begin{equation}
pe(I_{a},I_{b})=\frac{\alpha}{2}(1-SSIM(I_{a},I_{b}))+(1-\alpha)\left\Vert I_{a}-I_{b}\right\Vert _{1}.\label{eq:important-7}
\end{equation}

To ensure proper depth regularization in areas lacking texture, we
employ an edge-aware smooth loss, applied in the following manner:
\begin{equation}
L_{s}=\left|\partial_{x}d_{t}^{*}\right|e^{-\left|\partial_{x}I_{t}\right|}+\left|\partial_{y}d_{t}^{*}\right|e^{-\left|\partial_{y}I_{t}\right|}.\label{eq:important-8}
\end{equation}

\begin{table*}
\centering %
\begin{tabular}{|c|c|c|c|c|c|c|c|c|c|c|}
\hline 
Method & Train & Params & H$\times$W & AbsRel$\downarrow$ & SqRel$\downarrow$ & RMSE$\downarrow$ & RMSESlog$\downarrow$ & $\delta<1.25$$\uparrow$ & $\delta<1.25^{2}$$\uparrow$ & $\delta<1.25^{3}$$\uparrow$\tabularnewline
\hline 
\hline 
Struct2Depth 2 \citep{key-76} & M & - & 416$\times$128 & 0.145 & 1.737 & 7.280 & 0.205 & 0.813 & 0.942 & 0.976\tabularnewline
MonoDepth2 \citep{key-17} & MS & 34M & 416$\times$128 & 0.129 & 1.569 & 6.876 & 0.187 & 0.849 & 0.957 & 0.983\tabularnewline
ManyDepth \citep{key-60} & MS & 37M & 416$\times$128 & 0.114 & 1.193 & \textbf{6.223} & 0.170 & 0.875 & 0.967 & 0.989\tabularnewline
MambaDepth (Ours) & M & 30M & 416$\times$128 & \textbf{0.112} & \textbf{1.186} & 6.226 & \textbf{0.167} & \textbf{0.876} & \textbf{0.968} & \textbf{0.990}\tabularnewline
\hline 
\end{tabular}

\caption{\textbf{Cityscapes results.} Best results are marked in bold.}
\label{tab:Cityscapes results}
\end{table*}

\textbf{Masking Strategy. }In real-world settings, scenarios featuring
stationary cameras and moving objects can disrupt the usual assumptions
of a moving camera and static environment, negatively impacting the
performance of self-supervised depth estimators. Previous studies
have attempted to enhance depth prediction accuracy by incorporating
a motion mask, which addresses moving objects using scene-specific
instance segmentation models. However, this approach limits their
applicability to new, unencountered scenarios. To maintain scalability,
our method eschews the use of a motion mask for handling moving objects.
Instead, we adopt the auto-masking strategy outlined in \citep{key-17},
which filters out static pixels and areas of low texture that appear
unchanged between two consecutive frames in a sequence. This binary
mask $\mu$ is calculated as per \eqref{eq:important-9}, employing
the Iverson bracket notation:
\begin{equation}
\mu=[\underset{t'}{\mathrm{min}}\,pe(I_{t},I_{t'->t})<\underset{t'}{\mathrm{min}}\,pe(I_{t},I_{t'})].\label{eq:important-9}
\end{equation}

\textbf{Final Training Loss.} We formulate the final loss by combining
our per-pixel smooth loss with the masked photometric losses:
\begin{equation}
L=\mu L_{p}+\lambda L_{s}.\label{eq:important-10}
\end{equation}

\section{Experiments}

The effectiveness of MambaDepth is assessed using the public dataset
KITTI. We measure the model's performance using several established
metrics from \citep{key-9}. 

\subsection{Datasets and Experimental Protocol}

\textbf{KITTI} \citep{key-15} dataset, known for its stereo image
sequences, is widely utilized in self-supervised monocular depth estimation.
We employ the Eigen split \citep{key-9}, comprising about 26,000
images for training and 697 for testing. Our approach with MambaDepth
involves training it from the beginning on KITTI under minimal conditions:
it operates solely with auto-masking \citep{key-17}, without additional
stereo pairs or auxiliary data. For testing purposes, we maintain
a challenging scenario by using only a single frame as input, in contrast
to other methods that might use multiple frames to enhance accuracy.

\textbf{Cityscapes} \citep{key-6} dataset, noted for its complexity
and abundance of moving objects, serves as a testing ground to assess
the adaptability of MambaDepth. To this end, we conduct a zero-shot
evaluation on Cityscapes, utilizing a model pre-trained on KITTI.
It is crucial to highlight that, unlike many competing approaches,
we do not employ a motion mask in our evaluation. For data preparation,
we follow the same preprocessing procedures outlined in \citep{key-67},
which are also adopted by other baselines, converting the image sequences
into triplets. 

\textbf{Make3D} \citep{key-51}. To assess the capability of MambaDepth
to generalize to new, previously unseen images, the model, initially
trained on the KITTI dataset, was subjected to a zero-shot evaluation
using the Make3D dataset. Furthermore, supplementary visualizations
of depth maps are provided.

\subsection{Implementation Details}

We developed our model using the PyTorch framework \citep{key-44}.
It was trained on eight NVIDIA Tesla V100-SXM2 GPUs, with a batch
size of $8$. We pre-trained the architecture on the ImageNet-1k dataset,
subsequently using these pre-trained weights to initialize both the
encoder and decoder components of the model. In line with the approach
in \citep{key-17}, we applied color and flip augmentations to the
images during training. Both DepthNet and PoseNet were trained using
the Adam Optimizer \citep{key-32}, with $\beta_{1}$ set at 0.9 and
$\beta_{2}$ at 0.999, DepthNet being also trained concurrently. The
learning rate starts at $1e-4$ and reduces to $1e-5$ after $15$
epochs. We configured the SSIM weight at $\alpha=0.85$ and the weight
for the smooth loss term at $\lambda=1e-3$. 

\begin{figure}[th]
\centering \includegraphics[scale=0.14]{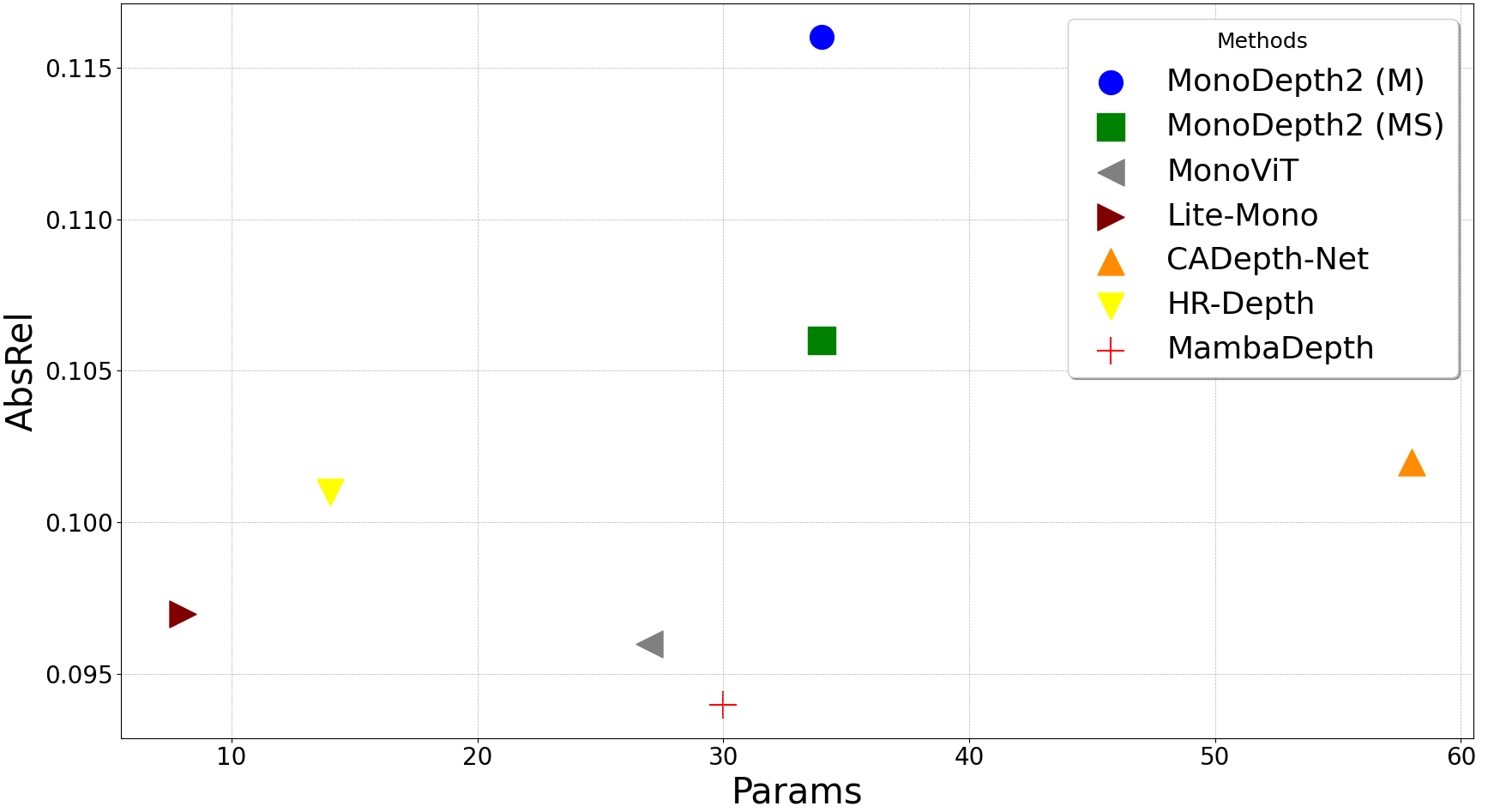}\label{fig:short-b-3-2-1-1}

\caption{Model parameters and performance with respect to the absolute relative
difference (Abs Rel) on the KITTI Eigen test set. Our model is more
efficient and accurate.}
\label{fig:Model parameters and performance with respect to the Abs Rel on the KITTI}
\end{figure}

\subsection{KITTI Results}

We evaluate our MambaDepth using the standard KITTI Eigen split \citep{key-9},
which consists of 697 images paired with raw LiDAR scans. Improved
ground truth labels \citep{key-55} are available for 652 of these
images. To address monocular scale ambiguity in depth models trained
on video sequences, we scale the estimated depth by the per-image
median ground truth \citep{key-67}.

Table \ref{tab:Performance comparison on KITTI eigen benchmark} presents
the results achieved by state-of-the-art self-supervised frameworks,
processing either low-resolution (640$\times$192) or high-resolution
(1024$\times$320) images. MambaDepth significantly outperforms existing
state-of-the-art methods across all training resolutions on all metrics,
some of them trained solely on stereo videos (\citep{key-59,key-45})
or with binocular videos while MambaDepth being trained entirely on
monocular videos. Notably, MambaDepth substantially surpasses MonoViT
\citep{key-65}, Lite-Mono \citep{key-73}, and MT-SfMLearner \citep{key-79},
the best contemporary attempts to use Transformers or self-attention
mechanism in self-supervised monocular depth estimation.

Table \ref{tab:Performance comparison using KITTI improved ground truth}
shows the same metrics computed using the improved ground truth labels
for images processed at 640$\times$192 resolution and 1024$\times$320
resolution. Again, MambaDepth consistently demonstrates higher accuracy.

Figure \ref{fig:Qualitative results on the KITTI eigen benchmark}
compares MambaDepth with some of its competitors, illustrating that
our model achieves significantly lower RMSE. This comparison highlights
MambaDepth's superior capability in modeling long-range relationships
between objects compared to existing models.

\begin{figure}[h]
\centering \includegraphics[scale=0.14]{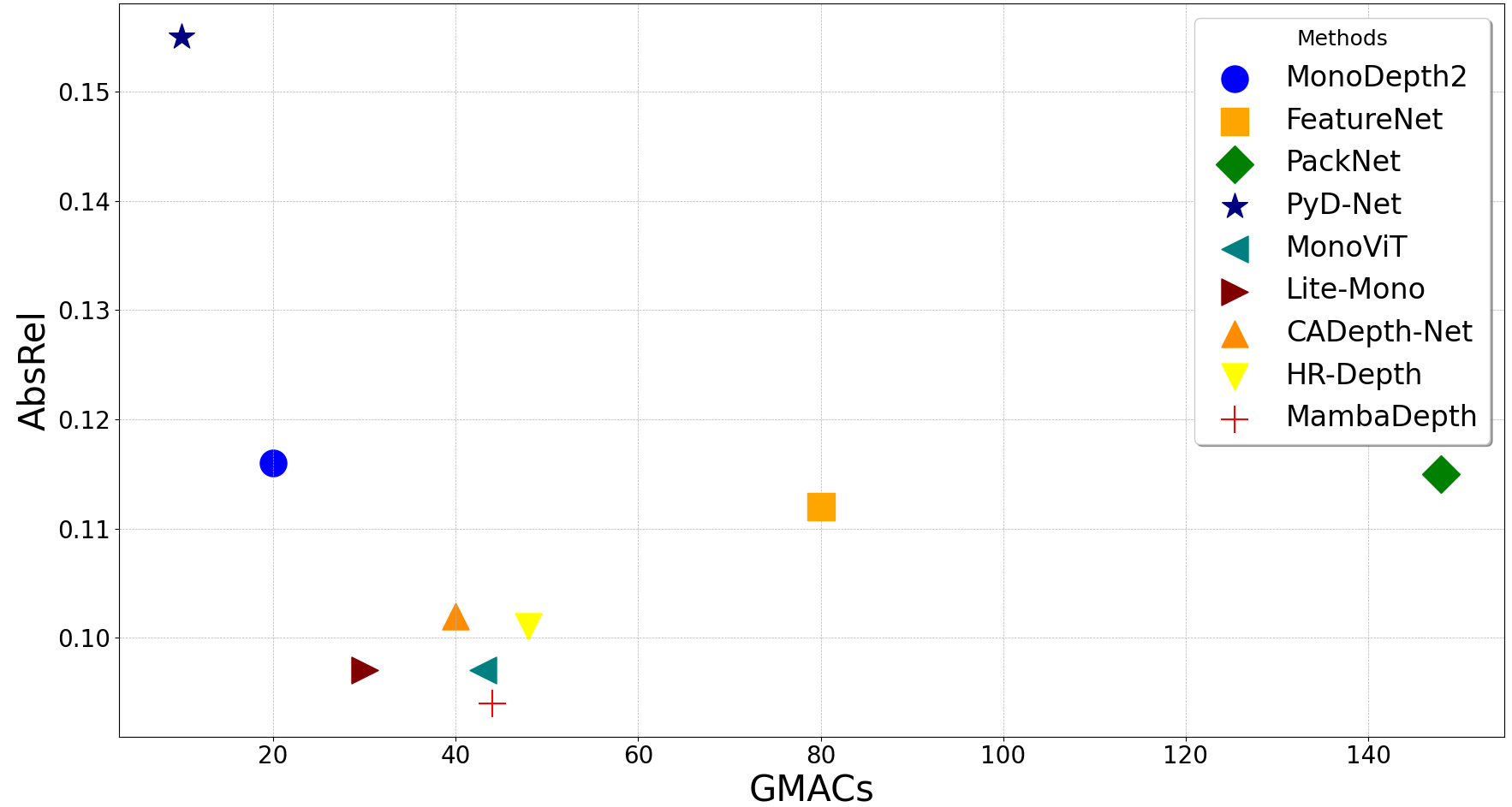}\label{fig:short-b-3-2-1}

\caption{Comparisons about computational costs. We compare AbsRel against Giga
Multiply-Add Caculation per Second (GMACs) on the KITTI Eigen test
set. Our model is more efficient and accurate.}
\label{fig:Comparisons about computational costs}
\end{figure}

Also the model is more efficient and accurate in terms of computations
costs, as depicted in Figure \ref{fig:Comparisons about computational costs},
where we compare the absolute relative difference against Giga Multiply-Add
Calculations per Second on the KITTI Eigen test set for more existing
state-of-the-art methods. As for parameters comparison, details are
present in Figure \ref{fig:Model parameters and performance with respect to the Abs Rel on the KITTI}.

\subsection{Cityscapes Results}

To evaluate the generalization of MambaDepth, we conducted zero-shot
evaluation. For this, we utilized the model pretrained on KITTI for
images processed at 416$\times$128 resolution. The results, summarized
in Table \ref{tab:Cityscapes results}, indicate that MambaDepth performs
exceptionally well being entirely trained with monocular videos, unlike
most baselines in Table \ref{tab:Cityscapes results} trained with
a combination of monocular videos and stereo pairs. Notably, MambaDepth
achieves a 1.75\% error reduction compared to the well-known ManyDepth
\citep{key-60}, which utilizes two frames (the previous and current)
as input. These findings underscore the superior generalization ability
of MambaDepth.

\subsection{Make3D Results}

\begin{figure}[h]
\centering\includegraphics[scale=0.12]{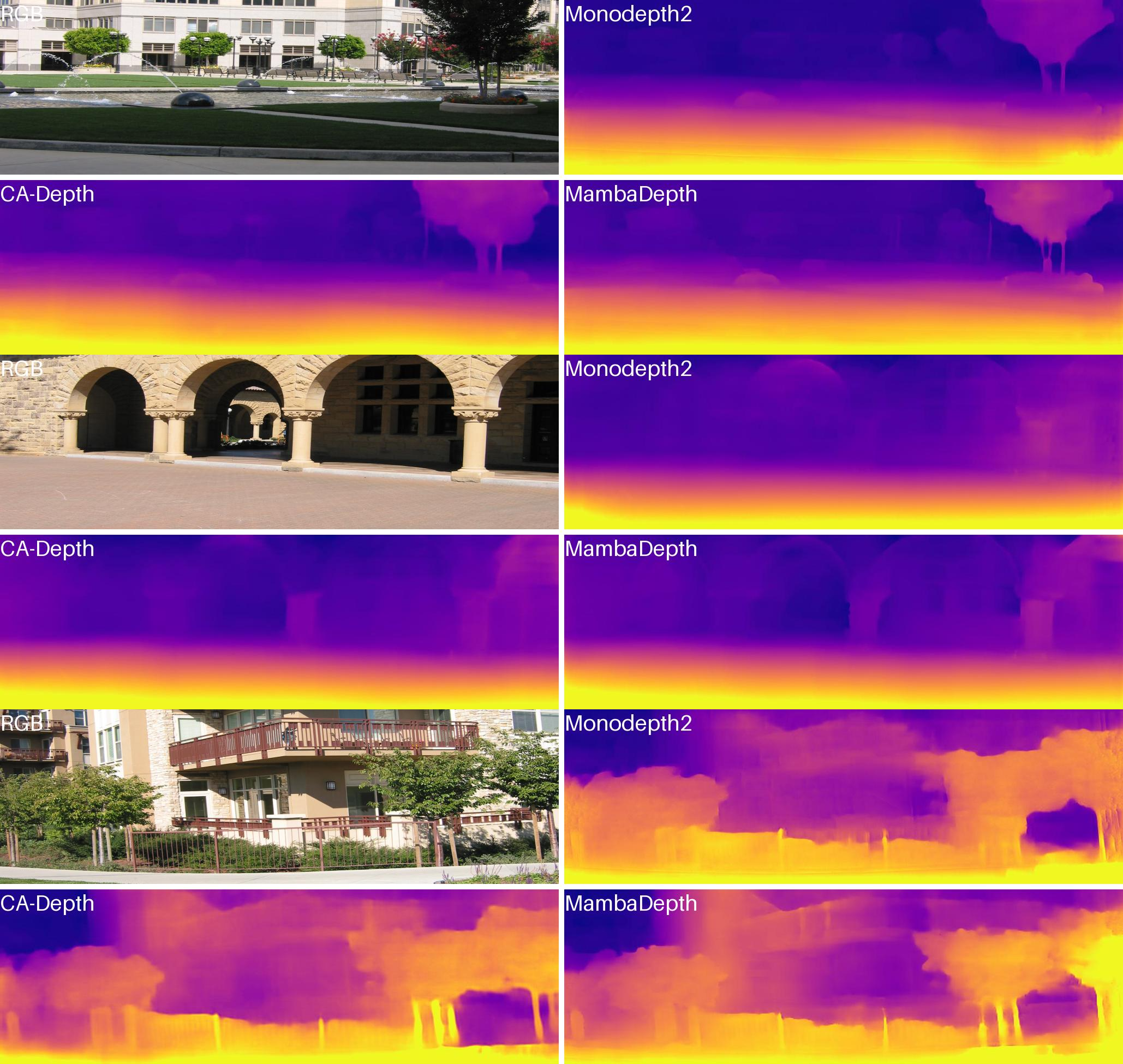}\label{fig:short-b-2-1-2}

\caption{\textbf{Qualitative Make3D results (Zero-shot). }Predictions by CA-Depth
\citep{key-74}, MonoDepth2 \citep{key-17} and our MambaDepth.\textbf{ }}
\label{fig:Qualitative Make3D results}
\end{figure}

To further assess the generalization capability of MambaDepth, we
conducted a zero-shot evaluation on the Make3D dataset \citep{key-51}
using the pretrained weights from KITTI. Adhering to the evaluation
protocol described in \citep{key-16}, we tested on a center crop
with a 2:1 aspect ratio. As illustrated in Table \ref{tab:Make3D results}
and Figure \ref{fig:Qualitative Make3D results}, MambaDepth outperformed
the baselines, delivering sharp depth maps with more precise scene
details. These results highlight the exceptional zero-shot generalization
ability of our model.

\begin{table}[h]
\centering %
\begin{tabular}{|c|c|c|c|c|}
\hline 
{\scriptsize{}Method} & {\scriptsize{}AbsRel $\downarrow$} & {\scriptsize{}SqRel $\downarrow$} & {\scriptsize{}RMSE $\downarrow$} & {\scriptsize{}RMSESlog $\downarrow$}\tabularnewline
\hline 
\hline 
{\scriptsize{}Zhou \citep{key-67}} & {\scriptsize{}0.383} & {\scriptsize{}5.321} & {\scriptsize{}10.470} & {\scriptsize{}0.478}\tabularnewline
{\scriptsize{}DDVO \citep{key-75}} & {\scriptsize{}0.387} & {\scriptsize{}4.720} & {\scriptsize{}8.090} & {\scriptsize{}0.204}\tabularnewline
{\scriptsize{}MonoDepth2 \citep{key-17}} & {\scriptsize{}0.322} & {\scriptsize{}3.589} & {\scriptsize{}7.417} & {\scriptsize{}0.163}\tabularnewline
{\scriptsize{}CADepth-Net \citep{key-74}} & {\scriptsize{}0.312} & {\scriptsize{}3.086} & {\scriptsize{}7.066} & {\scriptsize{}0.159}\tabularnewline
{\scriptsize{}MambaDepth (Ours)} & \textbf{\scriptsize{}0.307} & \textbf{\scriptsize{}2.405} & \textbf{\scriptsize{}6.858} & \textbf{\scriptsize{}0.153}\tabularnewline
\hline 
\end{tabular}

\caption{\textbf{Make3D results. }Best results are marked in bold.}
\label{tab:Make3D results}
\end{table}

\subsection{Ablation study}

Here, we explore the impact of initialization for MambaDepth using
KITTI dataset. We initialize MambaDepth with and without weights pretrained
on ImageNet. The results, displayed in Table \ref{tab:Ablation study on weights initialization},
suggest that stronger pretrained weights markedly improve MambaDepth's
subsequent effectiveness, highlighting the significant role these
initial weights play.

\begin{table}[h]
\centering %
\begin{tabular}{|c|c|c|c|c|}
\hline 
{\footnotesize{}Weights initialization} & {\footnotesize{}AbsRel $\downarrow$} & {\footnotesize{}SqRel $\downarrow$} & {\footnotesize{}RMSE $\downarrow$} & {\footnotesize{}RMSESlog $\downarrow$}\tabularnewline
\hline 
\hline 
{\footnotesize{}Xavier} & {\footnotesize{}0.114} & {\footnotesize{}0.881} & {\footnotesize{}4.700} & {\footnotesize{}0.191}\tabularnewline
{\footnotesize{}ImageNet pretraining} & \textbf{\footnotesize{}0.095} & \textbf{\footnotesize{}0.634} & \textbf{\footnotesize{}3.402} & \textbf{\footnotesize{}0.169}\tabularnewline
\hline 
\end{tabular}

\caption{\textbf{Ablation study on weights initialization for MambaDepth on
KITTI eigen benchmark}. Best results are marked in bold.}
\label{tab:Ablation study on weights initialization}
\end{table}

\section{Conclusions}

In our study, we re-examine the challenges of self-supervised monocular
depth estimation and present a novel and efficient approach, which
we have named MambaDepth. This method is designed to overcome the
challenges in capturing long-range dependencies, a limitation observed
in the localized nature of CNNs and the computational intensity of
Transformers. MambaDepth is a purely Mamba block-based U-Net-style
network for self-supervised monocular depth estimation. It achieves
outstanding state-of-the-art performance on the KITTI dataset. Additionally,
we showcase the enhanced generalizability of our model in various
settings. Our findings position MambaDepth as a leading contender
for future advanced self-supervised depth estimation networks.


\begin{thebibliography}{99}
\bibitem[1]{key-1}Shariq Farooq Bhat, Ibraheem Alhashim, and Peter
Wonka. Adabins: Depth estimation using adaptive bins. In \emph{Proceedings
of the IEEE/CVF Conference on Computer Vision and Pattern Recognition},
pages 4009--4018, 2021.

\bibitem[2]{key-2}Jiawang Bian, Zhichao Li, Naiyan Wang, Huangying
Zhan, Chunhua Shen, Ming-Ming Cheng, and Ian Reid. Unsupervised scale-consistent
depth and ego-motion learning from monocular video. \textit{Advances
in neural information processing systems}, 32, 2019.

\bibitem[2]{key-3}Vincent Casser, Soeren Pirk, Reza Mahjourian, and
Anelia Angelova. Unsupervised monocular depth and ego-motion learning
with structure and semantics. In \emph{Proceedings of the IEEE/CVF
Conference on Computer Vision and Pattern Recognition Workshops},
pages 0--0, 2019.

\bibitem{key-4}Po-Yi Chen, Alexander H Liu, Yen-Cheng Liu, and Yu-
Chiang Frank Wang. Towards scene understanding: Unsupervised monocular
depth estimation with semantic-aware representation. In \emph{Proceedings
of the IEEE/CVF Conference on computer vision and pattern recognition},
pages 2624--2632, 2019.

\bibitem{key-5}Yuhua Chen, Cordelia Schmid, and Cristian Sminchisescu.
Self-supervised learning with geometric constraints in monocular video:
Connecting flow, depth, and camera. In \emph{Proceedings of the IEEE/CVF
International Conference on Computer Vision}, pages 7063--7072, 2019.

\bibitem{key-6}Marius Cordts, Mohamed Omran, Sebastian Ramos, Timo
Rehfeld, Markus Enzweiler, Rodrigo Benenson, Uwe Franke, Stefan Roth,
and Bernt Schiele. The cityscapes dataset for semantic urban scene
understanding. In \emph{Proceedings of the IEEE conference on computer
vision and pattern recognition}, pages 3213--3223, 2016.

\bibitem{key-7}Raul Diaz and Amit Marathe. Soft labels for ordinal
regression. In \emph{Proceedings of the IEEE/CVF conference on computer
vision and pattern recognition}, pages 4738--4747, 2019.

\bibitem{key-8}Alexey Dosovitskiy, Lucas Beyer, Alexander Kolesnikov,
Dirk Weissenborn, Xiaohua Zhai, Thomas Unterthiner, Mostafa Dehghani,
Matthias Minderer, Georg Heigold, Sylvain Gelly, et al. An image is
worth 16x16 words: Transformers for image recognition at scale. \emph{arXiv
preprint arXiv:2010.11929}, 2020.

\bibitem{key-9}David Eigen and Rob Fergus. Predicting depth, surface
normals and semantic labels with a common multi-scale convolutional
architecture. In \emph{Proceedings of the IEEE international conference
on computer vision}, pages 2650--2658, 2015.

\bibitem{key-10}David Eigen, Christian Puhrsch, and Rob Fergus. Depth
map prediction from a single image using a multi-scale deep network.
\emph{Advances in neural information processing systems}, 27, 2014.

\bibitem{key-11}Ziyue Feng, Liang Yang, Longlong Jing, Haiyan Wang,
YingLi Tian, and Bing Li. Disentangling object motion and occlusion
for unsupervised multi-frame monocular depth. In \emph{European Conference
on Computer Vision}, pages 228--244. Springer, 2022.

\bibitem{key-12}Huan Fu, Mingming Gong, Chaohui Wang, Kayhan Bat-
manghelich, and Dacheng Tao. Deep ordinal regression network for monocular
depth estimation. In \emph{Proceedings of the IEEE conference on computer
vision and pattern recognition}, pages 2002--2011, 2018.

\bibitem{key-13}Divyansh Garg, Yan Wang, Bharath Hariharan, Mark
Campbell, Kilian Q Weinberger, and Wei-Lun Chao. Wasserstein distances
for stereo disparity estimation. \emph{Advances in Neural Information
Processing Systems}, 33:22517--22529, 2020.

\bibitem{key-14}Ravi Garg, Vijay Kumar Bg, Gustavo Carneiro, and
Ian Reid. Unsupervised cnn for single view depth estimation: Geometry
to the rescue. In \emph{Computer Vision--ECCV 2016: 14th European
Conference, Amsterdam, The Netherlands, October 11-14, 2016, Proceedings,
Part VIII 14}, pages 740--756. Springer, 2016.

\bibitem{key-15}Andreas Geiger, Philip Lenz, Christoph Stiller, and
Raquel Urtasun. Vision meets robotics: The kitti dataset. \emph{The
International Journal of Robotics Research}, 32(11):1231--1237, 2013.

\bibitem{key-16}Clément Godard, Oisin Mac Aodha, and Gabriel J Brostow.
Unsupervised monocular depth estimation with left-right consistency.
In \emph{Proceedings of the IEEE conference on computer vision and
pattern recognition}, pages 270--279, 2017.

\bibitem{key-17}Clément Godard, Oisin Mac Aodha, Michael Firman,
and Gabriel J Brostow. Digging into self-supervised monocular depth
estimation. In \emph{Proceedings of the IEEE/CVF international conference
on computer vision}, pages 3828--3838, 2019.

\bibitem{key-18}Juan Luis Gonzalez and Munchurl Kim. Plade-net: Towards
pixel-level accuracy for self-supervised single-view depth estimation
with neural positional encoding and distilled matting loss. In \emph{Proceedings
of the IEEE/CVF Conference on Computer Vision and Pattern Recognition},
pages 6851--6860, 2021.

\bibitem{key-19}Juan Luis GonzalezBello and Munchurl Kim. Forget
about the lidar: Self-supervised depth estimators with med probability
volumes. \emph{Advances in Neural Information Processing Systems},
33:12626--12637, 2020.

\bibitem{key-20}Ariel Gordon, Hanhan Li, Rico Jonschkowski, and Anelia
Angelova. Depth from videos in the wild: Unsupervised monocular depth
learning from unknown cameras. In \emph{Proceedings of the IEEE/CVF
International Conference on Computer Vision}, pages 8977--8986, 2019.

\bibitem{key-21}Albert Gu. \emph{Modeling Sequences with Structured
State Spaces}. PhD thesis, Stanford University, 2023.

\bibitem{key-22}Albert Gu and Tri Dao. Mamba: Linear-time sequence
modeling with selective state spaces. \emph{arXiv preprint arXiv:2312.00752},
2023.

\bibitem{key-23}Albert Gu, Tri Dao, Stefano Ermon, Atri Rudra, and
Christopher Ré. Hippo: Recurrent memory with optimal polynomial projections.
\emph{Advances in neural information processing systems}, 33:1474--1487,
2020.

\bibitem{key-24}Albert Gu, Karan Goel, and Christopher Ré. Efficiently
modeling long sequences with structured state spaces. \emph{arXiv
preprint arXiv:2111.00396}, 2021.

\bibitem{key-25}Albert Gu, Isys Johnson, Karan Goel, Khaled Saab,
Tri Dao, Atri Rudra, and Christopher Ré. Combining recurrent, convolutional,
and continuous-time models with linear state space layers. \emph{Advances
in neural information processing systems, 34:572--585}, 2021.

\bibitem{key-26}Vitor Guizilini, Rui Hou, Jie Li, Rares Ambrus, and
Adrien Gaidon. Semantically-guided representation learning for self-supervised
monocular depth. \emph{arXiv preprint arXiv:2002.12319}, 2020.

\bibitem{key-27}Xiaoyang Guo, Hongsheng Li, Shuai Yi, Jimmy Ren,
and Xiaogang Wang. Learning monocular depth by distilling cross-domain
stereo networks. In \emph{Proceedings of the European conference on
computer vision (ECCV)}, pages 484-- 500, 2018.

\bibitem{key-28}Lam Huynh, Phong Nguyen-Ha, Jiri Matas, Esa Rahtu,
and Janne Heikkilä. Guiding monocular depth estimation using depth-attention
volume. In \emph{Computer Vision--ECCV 2020: 16th European Conference,
Glasgow, UK, August 23--28, 2020, Proceedings, Part XXVI 16}, pages
581--597. Springer, 2020.

\bibitem{key-29}Md Mohaiminul Islam and Gedas Bertasius. Long movie
clip classification with state-space video models. In \emph{European
Conference on Computer Vision}, pages 87--104. Springer, 2022.

\bibitem{key-30}Max Jaderberg, Karen Simonyan, Andrew Zisserman,
et al. Spatial transformer networks. \emph{Advances in neural information
processing systems}, 28, 2015.

\bibitem{key-31}Adrian Johnston and Gustavo Carneiro. Self-supervised
monocular trained depth estimation using self-attention and discrete
disparity volume. In \emph{Proceedings of the ieee/cvf conference
on computer vision and pattern recognition}, pages 4756--4765, 2020.

\bibitem{key-32}Diederik P Kingma and Jimmy Ba. Adam: A method for
stochastic optimization. \emph{arXiv preprint arXiv:1412.6980}, 2014.

\bibitem{key-33}Marvin Klingner, Jan-Aike Termöhlen, Jonas Mikolajczyk,
and Tim Fingscheidt. Self-supervised monocular depth estimation: Solving
the dynamic object problem by semantic guidance. In \emph{Computer
Vision--ECCV 2020: 16th European Conference, Glasgow, UK, August
23--28, 2020, Proceedings, Part XX 16}, pages 582--600. Springer,
2020.

\bibitem{key-34}Maria Klodt and Andrea Vedaldi. Supervising the new
with the old: learning sfm from sfm. In \emph{Proceedings of the European
conference on computer vision (ECCV)}, pages 698-- 713, 2018.

\bibitem{key-35}Yevhen Kuznietsov, Jorg Stuckler, and Bastian Leibe.
Semi-supervised deep learning for monocular depth map prediction.
In \emph{Proceedings of the IEEE conference on computer vision and
pattern recognition}, pages 6647--6655, 2017.

\bibitem{key-36}Yann LeCun, Yoshua Bengio, et al. Convolutional networks
for images, speech, and time series. \emph{The handbook of brain theory
and neural networks}, 3361(10):1995, 1995.

\bibitem{key-37}Hanhan Li, Ariel Gordon, Hang Zhao, Vincent Casser,
and Anelia Angelova. Unsupervised monocular depth learning in dynamic
scenes. In \emph{Conference on Robot Learning}, pages 1908--1917.
PMLR, 2021.

\bibitem{key-38}Zhenyu Li, Zehui Chen, Xianming Liu, and Junjun Jiang.
Depthformer: Exploiting long-range correlation and local information
for accurate monocular depth estimation. \emph{arXiv preprint arXiv:2203.14211},
2022.

\bibitem{key-39}Yue Liu, Yunjie Tian, Yuzhong Zhao, Hongtian Yu,
Lingxi Xie, Yaowei Wang, Qixiang Ye, and Yunfan Liu. Vmamba: Visual
state space model. \emph{arXiv preprint arXiv:2401.10166}, 2024.

\bibitem{key-40}Ze Liu, Yutong Lin, Yue Cao, Han Hu, Yixuan Wei,
Zheng Zhang, Stephen Lin, and Baining Guo. Swin transformer: Hierarchical
vision transformer using shifted windows. In \emph{Proceedings of
the IEEE/CVF international conference on computer vision}, pages 10012--10022,
2021.

\bibitem{key-41}Yue Luo, Jimmy Ren, Mude Lin, Jiahao Pang, Wenxiu
Sun, Hongsheng Li, and Liang Lin. Single view stereo matching. In
\emph{Proceedings of the IEEE Conference on Computer Vision and Pattern
Recognition}, pages 155--163, 2018.

\bibitem{key-42}Jun Ma, Feifei Li, and Bo Wang. U-mamba: Enhancing
long-range dependency for biomedical image segmentation. \emph{arXiv
preprint arXiv:2401.04722}, 2024.

\bibitem{key-43}Eric Nguyen, Karan Goel, Albert Gu, Gordon Downs,
Preey Shah, Tri Dao, Stephen Baccus, and Christopher Ré. S4nd: Modeling
images and videos as multidimensional signals with state spaces. \emph{Advances
in neural information processing systems}, 35:2846--2861, 2022.

\bibitem{key-44}Adam Paszke, Sam Gross, Francisco Massa, Adam Lerer,
James Bradbury, Gregory Chanan, Trevor Killeen, Zeming Lin, Natalia
Gimelshein, Luca Antiga, et al. Pytorch: An imperative style, high-performance
deep learning library. \emph{Advances in neural information processing
systems}, 32, 2019.

\bibitem{key-45}Rui Peng, Ronggang Wang, Yawen Lai, Luyang Tang,
and Yangang Cai. Excavating the potential capacity of self-supervised
monocular depth estimation. In \emph{Proceedings of the IEEE/CVF International
Conference on Computer Vision}, pages 15560--15569, 2021.

\bibitem{key-46}Sudeep Pillai, Rareş Ambruş, and Adrien Gaidon. Superdepth:
Self-supervised, super-resolved monocular depth estimation. In \emph{2019
International Conference on Robotics and Automation (ICRA)}, pages
9250--9256. IEEE, 2019.

\bibitem{key-47}Andrea Pilzer, Stephane Lathuiliere, Nicu Sebe, and
Elisa Ricci. Refine and distill: Exploiting cycle-inconsistency and
knowledge distillation for unsupervised monocular depth estimation.
In \emph{Proceedings of the IEEE/CVF Conference on Computer Vision
and Pattern Recognition}, pages 9768-- 9777, 2019.

\bibitem{key-48}René Ranftl, Alexey Bochkovskiy, and Vladlen Koltun.
Vision transformers for dense prediction. In \emph{Proceedings of
the IEEE/CVF international conference on computer vision}, pages 12179--12188,
2021.

\bibitem{key-49}René Ranftl, Katrin Lasinger, David Hafner, Konrad
Schindler, and Vladlen Koltun. Towards robust monocular depth estimation:
Mixing datasets for zero-shot cross-dataset transfer. \emph{IEEE transactions
on pattern analysis and machine intelligence}, 44(3):1623--1637,
2020.

\bibitem{key-50}Anurag Ranjan, Varun Jampani, Lukas Balles, Kihwan
Kim, Deqing Sun, Jonas Wulff, and Michael J Black. Competitive collaboration:
Joint unsupervised learning of depth, camera motion, optical flow
and motion segmentation. In \emph{Proceedings of the IEEE/CVF conference
on computer vision and pattern recognition}, pages 12240--12249,
2019.

\bibitem{key-51}Ashutosh Saxena, Min Sun, and Andrew Y Ng. Learning
3-d scene structure from a single still image. In \emph{2007 IEEE
11th international conference on computer vision}, pages 1-- 8. IEEE,
2007.

\bibitem{key-52}Daniel Scharstein and Richard Szeliski. A taxonomy
and evaluation of dense two-frame stereo correspondence algorithms.
\emph{International journal of computer vision}, 47:7--42, 2002.

\bibitem{key-53}Chang Shu, Kun Yu, Zhixiang Duan, and Kuiyuan Yang.
Feature-metric loss for self-supervised learning of depth and egomotion.
In \emph{European Conference on Computer Vision}, pages 572--588.
Springer, 2020.

\bibitem{key-54}Yi Tay, Mostafa Dehghani, Samira Abnar, Yikang Shen,
Dara Bahri, Philip Pham, Jinfeng Rao, Liu Yang, Sebastian Ruder, and
Donald Metzler. Long range arena: A benchmark for efficient transformers.
\emph{arXiv preprint arXiv:2011.04006}, 2020.

\bibitem{key-55}Jonas Uhrig, Nick Schneider, Lukas Schneider, Uwe
Franke, Thomas Brox, and Andreas Geiger. Sparsity invariant cnns.
In \emph{2017 international conference on 3D Vision (3DV)}, pages
11--20. IEEE, 2017.

\bibitem{key-56}Ashish Vaswani, Noam Shazeer, Niki Parmar, Jakob
Uszkoreit, Llion Jones, Aidan N Gomez, Łukasz Kaiser, and Illia Polosukhin.
Attention is all you need. \emph{Advances in neural information processing
systems}, 30, 2017.

\bibitem{key-57}Sudheendra Vijayanarasimhan, Susanna Ricco, Cordelia
Schmid, Rahul Sukthankar, and Katerina Fragkiadaki. Sfm-net: Learning
of structure and motion from video. \emph{arXiv preprint arXiv:1704.07804},
2017.

\bibitem{key-58}Youhong Wang, Yunji Liang, Hao Xu, Shaohui Jiao,
and Hongkai Yu. Sqldepth: Generalizable self-supervised fine-structured
monocular depth estimation. \emph{arXiv preprint arXiv:2309.00526},
2023.

\bibitem{key-59}Jamie Watson, Michael Firman, Gabriel J Brostow,
and Daniyar Turmukhambetov. Self-supervised monocular depth hints.
In \emph{Proceedings of the IEEE/CVF International Conference on Computer
Vision}, pages 2162--2171, 2019.

\bibitem{key-60}Jamie Watson, Oisin Mac Aodha, Victor Prisacariu,
Gabriel Brostow, and Michael Firman. The temporal opportunist: Self-supervised
multi-frame monocular depth. In \emph{Proceedings of the IEEE/CVF
Conference on Computer Vision and Pattern Recognition}, pages 1164--1174,
2021.

\bibitem{key-61}Guanglei Yang, Hao Tang, Mingli Ding, Nicu Sebe,
and Elisa Ricci. Transformer-based attention networks for continuous
pixel-wise prediction. In \emph{Proceedings of the IEEE/CVF International
Conference on Computer vision}, pages 16269--16279, 2021.

\bibitem{key-62}Zhenheng Yang, Peng Wang, Wei Xu, Liang Zhao, and
Ramakant Nevatia. Unsupervised learning of geometry from videos with
edge-aware depth-normal consistency. In \emph{Proceedings of the AAAI
Conference on Artificial Intelligence}, volume 32, 2018.

\bibitem{key-63}Zhichao Yin and Jianping Shi. Geonet: Unsupervised
learning of dense depth, optical flow and camera pose. In \emph{Proceedings
of the IEEE conference on computer vision and pattern recognition},
pages 1983--1992, 2018.

\bibitem{key-64}Huangying Zhan, Ravi Garg, Chamara Saroj Weerasekera,
Kejie Li, Harsh Agarwal, and Ian Reid. Unsupervised learning of monocular
depth estimation and visual odometry with deep feature reconstruction.
In \emph{Proceedings of the IEEE conference on computer vision and
pattern recognition}, pages 340--349, 2018.

\bibitem{key-65}Chaoqiang Zhao, Youmin Zhang, Matteo Poggi, Fabio
Tosi, Xianda Guo, Zheng Zhu, Guan Huang, Yang Tang, and Stefano Mattoccia.
Monovit: Self-supervised monocular depth estimation with a vision
transformer. In \emph{2022 International Conference on 3D Vision (3DV)},
pages 668--678. IEEE, 2022.

\bibitem{key-66}Jiawei Zhao, Ke Yan, Yifan Zhao, Xiaowei Guo, Feiyue
Huang, and Jia Li. Transformer-based dual relation graph for multi-label
image recognition. In \emph{Proceedings of the IEEE/CVF international
conference on computer vision}, pages 163--172, 2021.

\bibitem{key-67}Tinghui Zhou, Matthew Brown, Noah Snavely, and David
G Lowe. Unsupervised learning of depth and ego-motion from video.
In \emph{Proceedings of the IEEE conference on computer vision and
pattern recognition}, pages 1851--1858, 2017.

\bibitem{key-68}Lianghui Zhu, Bencheng Liao, Qian Zhang, Xinlong
Wang, Wenyu Liu, and Xinggang Wang. Vision mamba: Efficient visual
representation learning with bidirectional state space model. \emph{arXiv
preprint arXiv:2401.09417}, 2024.

\bibitem{key-69}Shengjie Zhu, Garrick Brazil, and Xiaoming Liu. The
edge of depth: Explicit constraints between segmentation and depth.
In \emph{Proceedings of the IEEE/CVF conference on computer vision
and pattern recognition}, pages 13116--13125, 2020.

\bibitem{key-72}Lyu, Xiaoyang, Liang Liu, Mengmeng Wang, Xin Kong,
Lina Liu, Yong Liu, Xinxin Chen, and Yi Yuan. Hr-depth: High resolution
self-supervised monocular depth estimation. In \emph{Proceedings of
the AAAI conference on artificial intelligence}, vol. 35, no. 3, pp.
2294-2301. 2021.

\bibitem{key-73}Zhang, Ning, Francesco Nex, George Vosselman, and
Norman Kerle. Lite-mono: A lightweight cnn and transformer architecture
for self-supervised monocular depth estimation. In \emph{Proceedings
of the IEEE/CVF Conference on Computer Vision and Pattern Recognition},
pp. 18537-18546. 2023.

\bibitem{key-74}Yan, Jiaxing, Hong Zhao, Penghui Bu, and YuSheng
Jin. Channel-wise attention-based network for self-supervised monocular
depth estimation. In \emph{2021 International Conference on 3D vision
(3DV)}, pp. 464-473. IEEE, 2021.

\bibitem{key-75}Chaoyang Wang, José Miguel Buenaposada, Rui Zhu,
and Simon Lucey. Learning depth from monocular videos using direct
methods. In \emph{Proceedings of the IEEE conference on computer vision
and pattern recognition}, pages 2022--2030, 2018.

\bibitem{key-76}Vincent Casser, Soeren Pirk, Reza Mahjourian, and
Anelia Angelova. Unsupervised monocular depth and ego-motion learning
with structure and semantics. In \emph{Proceedings of the IEEE/CVF
Conference on Computer Vision and Pattern Recognition Workshops},
pages 0--0, 2019.

\bibitem{key-77}Hang Zhou, David Greenwood, and Sarah Taylor. Self-supervised
monocular depth estimation with internal feature fusion. \emph{arXiv
preprint arXiv:2110.09482}, 2021.

\bibitem{key-78}Vitor Guizilini, Rares Ambrus, Sudeep Pillai, and
Adrien Gaidon. Packnet-sfm: 3d packing for self-supervised monocular
depth estimation. \emph{arXiv preprint arXiv:1905.02693}, 5(1), 2019.

\bibitem{key-79}Arnav Varma., Hemang Chawla., Bahram Zonooz., and
Elahe Arani. Transformers in self-supervised monocular depth estimation
with unknown camera intrinsics. In \emph{Proceedings of the 17th International
Joint Conference on Computer Vision, Imaging and Computer Graphics
Theory and Applications} - Volume 4: VISAPP, pages 758--769. IN-
STICC, SciTePress, 2022.{]}
\end{thebibliography}
\end{document}